\documentclass[conference]{IEEEtran}
\IEEEoverridecommandlockouts
\usepackage{cite}
\usepackage{amsmath,amssymb,amsfonts}
\usepackage{algorithmic}
\usepackage{graphicx}
\usepackage{textcomp}
\usepackage{xcolor}
\usepackage{algorithm}
\usepackage{multirow}
\usepackage[utf8]{inputenc}
\usepackage{url}
\usepackage{color}

\def\BibTeX{{\rm B\kern-.05em{\sc i\kern-.025em b}\kern-.08em
    T\kern-.1667em\lower.7ex\hbox{E}\kern-.125emX}}
\begin{document}

\title{An underwater binocular stereo matching algorithm based on the best search domain\\}

\author{\IEEEauthorblockN{Yimin Peng, Zijing Fang,  Yunlong Li}}
\maketitle

\begin{abstract}
Binocular stereo vision is an important branch of machine vision, which imitates the human eye and matches the left and right images captured by the camera based on epipolar constraints. The matched disparity map can be calculated according to the camera imaging model to obtain a depth map, and then the depth map is converted to a point cloud image to obtain spatial point coordinates, thereby achieving the purpose of ranging. However, due to the influence of illumination under water, the captured images no longer meet the epipolar constraints, and the changes in imaging models make traditional calibration methods no longer applicable. Therefore, this paper proposes a new underwater real-time calibration method and a matching method based on the best search domain to improve the accuracy of underwater distance measurement using binoculars.
\end{abstract}


\section{Introduction}
Robots must "see clearly" underwater, machine vision is the most important thing, and the camera is the "eyes" of the robot, and its calibration is the basis of machine vision. The edge part of the image obtained by the camera will be distorted. After calibration, the internal and external parameters of the camera and the distortion parameters can be obtained to obtain the correct image. Most of the most advanced underwater vision systems are manually calibrated in shallow water and can be used in the high seas without modification. However, the underwater situation is complex, and the refractive index of the water changes adaptively according to salinity, temperature, depth or other underwater environment indicators. At this time, the underwater vision system is prone to large errors. Therefore, in order to avoid the influence of nonlinear geometric transformation caused by underwater light refraction, this project is based on the machine learning method and the imu sensor to establish an underwater model of water depth, temperature, and refractive index transformation, which is analogous and equivalent to the calibration model on land , To achieve underwater real-time calibration. Generally, an IMU contains three single-axis accelerometers and three single-axis gyroscopes. The accelerometer detects the acceleration signal of the object in the independent three-axis coordinate system of the carrier, and the gyroscope detects the angular velocity signal of the carrier relative to the navigation coordinate system. The angular velocity and acceleration of the object in three-dimensional space, and the posture of the object is calculated based on this. It has very important application value in navigation. Based on the machine learning method, learn the correspondence between the two-dimensional observation and the three-dimensional posture from the training samples obtained in advance under different postures, and apply the learned decision rules or regression functions to the samples, and the result is used as the posture of the sample estimate. Learning-based methods generally use global observation features, do not need to detect or recognize local features of objects, and have good robustness.

\section{Problem description}
Binocular vision technology is generally divided into five steps: image acquisition-image preprocessing-camera calibration-stereo matching-three-dimensional reconstruction. In this paper, a sealed cabin is used to encapsulate and stabilize the binocular camera to achieve the purpose of waterproofing. The arranged environment is taken underwater to obtain images. Since the vision system needs to work in an underwater environment, during the underwater imaging process, light must pass through water/glass/air/lens (they are a variety of different media) and finally reach the imaging plane for imaging; therefore, such as As shown in Figure 1, when light passes through these different types of media, refraction occurs.
Most of the most advanced underwater vision systems are manually calibrated in shallow water and can be used in the high seas without modification. However, the underwater situation is complicated, and the refractive index of the water changes adaptively according to salinity, temperature, depth or other underwater environment indicators. At this time, the underwater vision system is prone to large errors. Therefore, in order to avoid the influence of nonlinear geometric transformation caused by underwater light refraction, this project is based on the machine learning method and the imu sensor to establish an underwater model of water depth, temperature, and refractive index transformation, which is analogous and equivalent to the calibration model on land , To achieve underwater real-time calibration. Generally, an IMU contains three single-axis accelerometers and three single-axis gyroscopes. The accelerometer detects the acceleration signal of an object in the independent three-axis coordinate system of the carrier, while the gyroscope detects the angular velocity signal of the carrier relative to the navigation coordinate system. The angular velocity and acceleration of the object in three-dimensional space, and the posture of the object is calculated based on this. It has very important application value in navigation.

\section{Establishment of underwater binocular vision model}
In the land-based binocular stereo vision system, the target object and the camera are all located in the air, but in the stereo vision system in the underwater environment involved in this article, the above two are located in water and air respectively, so the light will pass through different media, and finally Image on the image plane. During the experiment, this article stabilized the binocular camera in the airtight cabin, fixed the airtight cabin with clamps, so that it was immersed in water to take pictures of the target object.

The optical axes of the left and right cameras are placed parallel to the outside of the waterproof cover and perpendicular to the refraction plane. As shown in the figure, the distance between the optical center of the camera and the glass cover is  , the thickness of the transparent waterproof glass is  , and the water object point  passes through the waterproof cover. At  and point  , they are imaged as point  in a linear camera. However, in actual situations, due to equipment factors and the angle of view, the optical axes of the two cameras cannot be completely parallel and placed perpendicular to the refraction plane, and the cameras and the refraction plane will be inclined at a certain angle.

\end{document}